\documentclass[a4paper]{article}

\usepackage{INTERSPEECH2019}

\title{Curriculum-based transfer learning for an effective end-to-end spoken language understanding and domain portability}
\name{Antoine Caubri\`ere$^1$, Natalia Tomashenko$^2$, Antoine Laurent$^1$, Emmanuel Morin$^3$, \\
Nathalie Camelin$^1$, Yannick Est\`eve$^2$}
\address{
  $^1$LIUM - Le Mans University\\
  $^2$LIA - Avignon University\\
  $^3$LS2N - University of Nantes, CNRS}
  
\email{firstname.lastname@univ-lemans.fr, firstname.lastname@univ-avignon.fr, firstname.lastname@univ-nantes.fr}

\begin{document}

\maketitle
\begin{abstract}
We present an end-to-end approach to extract semantic concepts directly from the speech audio signal.
To overcome the lack of data available for this spoken language understanding approach, we investigate the use of a transfer learning strategy based on the principles of curriculum learning. This approach allows us to exploit out-of-domain data that can help to prepare a fully neural architecture.
Experiments are carried out on the French MEDIA and PORTMEDIA corpora and show that this end-to-end SLU approach reaches the best results ever published on this task.
We compare our approach to a classical pipeline approach that uses ASR, POS tagging, lemmatizer, chunker... and other NLP tools that aim to enrich ASR outputs that feed an SLU text to concepts system. 
Last, we explore the promising capacity of our end-to-end SLU approach to address the problem of domain portability.

\end{abstract}

\section{Introduction}

Thanks to great advances in automatic speech recognition (ASR) these last years, mainly due to advances on deep neural networks for both acoustic and language modelling, performance of spoken language understanding (SLU) systems has made notable progress. 
SLU is a term that refers to different NLP tasks applied to spoken language. 
For instance, this can be named entity extraction~\cite{kubala1998named}, call routing~\cite{gorin1997may}, domain classification at the utterance level~\cite{yaman2008integrative, tur2012towards}, at the conversation level~\cite{morchid2013theme}, \textsl{etc.}
Such SLU systems are usually natural language processing (NLP) systems applied to ASR outputs~\cite{tur2011spoken}, and better quality of automatic transcriptions leads to better performance of NLP systems.
In this study, the SLU targeted task is slot filling. 
This is an important task involved in goal-oriented human/machine dialogues~\cite{chen2013unsupervised}. 
Its goal consists on automatically extracting semantic concepts from a utterance, and on extracting values associated to these instances of concepts. 
For example, in the sentence "Interspeech 2019 will occur in Austria", a concept to extract could be COUNTRY-LOCATION and its value 'Austria'. 
In spoken dialogue systems, semantic slots are usually predefined in order to feed the semantic representation used by a dialogue manager module.
For slot filling task as for other SLU tasks, the classical approach uses a pipeline process: successive treatments are applied, from speech signal to concepts. 
First an ASR is applied to speech audio signal to produce automatic transcriptions. 
These transcriptions are processed by different NLP tools, like part-of-speech (POS) tagger, lemmatizer, chunker, semantic labeler... in order to enrich the ASR outputs.
Enriched transcriptions are then processed by the SLU tool that will extract both concepts and their values to fill semantic slots~\cite{simonnet2017asr}.

In this work, we focus on an end-to-end neural approach that extracts both concepts and values directly from speech. 
We recently presented a preliminary work on this approach that got promising results~\cite{ghannay2018end}.
A motivation to end-to-end approach for semantic extraction directly from speech is to limit the ASR error propagation and to take benefit from a joint optimization of ASR and SLU parts to the final task.
In this new study, we show how it is possible to reach state-of-the-art performance through a such end-to-end approach, that simplifies a lot the entire process in comparison to a classical pipeline approach.
We use the same neural architecture we presented in~\cite{ghannay2018end}, but we apply a curriculum-based transfer learning that strongly helps to get state-of-the-art performance.
Basic idea of curriculum learning is to \textsl{"start small, learn easier aspects of the task or easier sub tasks, and then gradually increase the difficulty level"}~\cite{bengio2009curriculum}. 
Usually this approach consists on ordering training samples without modifying the neural architecture during the training process.
We adapt this approach to design a sequence of transfer learning processes~\cite{bengio2011deep}, from a general task to the target specialized task.
While large amount of data is needed to train an SLU end-to-end neural model from speech, this approach allows us to deal with the lack of data related to the target task.
Last, we also investigate the capacity of domain portability brought by our approach, that consists on starting from an existing SLU model dedicated to a task, MEDIA~\cite{bonneau2005semantic}, in order to build a new SLU model dedicated to another task, PORT-MEDIA~\cite{lefevre2012leveraging}.

\section{Related work}
Thanks to the success of end-to-end ASR systems like Deep speech, the Baidu's system~\cite{hannun2014deep} that reached great performance on speech recognition through a fully neural architecture, some research teams recently investigated the use of end-to-end approaches for different tasks applied to speech. 
For instance, some studies explored end-to-end spoken language translation~\cite{berard2016listen, weiss2017sequence, berard2018end} showing that such approaches work, even if they currently do not reach state-of-the-art performance on this task~\cite{jan2018iwslt}.
In \cite{serdyuk2018towards}, authors proposed an end-to-end approach for SLU, for both speech-to-domain and speech-to-intent tasks. Even if they did not reach state-of-the-art performance, their results were promising. 
We shared the same conclusion in our previous work in~\cite{ghannay2018end}, that proved that an end-to-end neural approach for slot filling task was possible, without reaching state-of-art performances.

In this paper, we show how well we have taken a step by using the same neural architecture as the one we proposed in~\cite{ghannay2018end}.
Thanks to the use of a transfer learning approach inspired by curriculum learning~\cite{elman1993learning, bengio2009curriculum}, we are now able to reach state-of-the-art performance. A curriculum learning approach has also been recently proposed with success for machine translation in~\cite{platanios2019competence}.
At the end, we also address the issue of domain portability for SLU systems~\cite{lefevre2012leveraging,jabaian2012portability} that can obviously be tackled as a transfer learning problem.

\section{SLU end-to-end neural architecture}

\label{sec:neuralarch}

The SLU end-to-end neural architecture used in this study is the same as the one recently proposed in~\cite{ghannay2018end}. 
This architecture is largely inspired by Deep Speech~2~\cite{amodei2016deep}.
This deep neural network is composed of a stack of two 2D-invariant convolutional layers, followed by five bidirectional long short term  layers with sequence-wise batch normalization, a fully connected layer, and a final softmax layer. 
A spectrogram of power normalized audio clips calculated on 20ms windows is used as input features.



The system is trained end-to-end using the CTC loss function \cite{graves2006connectionist}, in order to predict a sequence of characters from the input audio. 
This sequence of characters represents words and semantic concepts.
Instead of applying the BIO approach as classically used to delimit semantic concepts on the words~\cite{hahn2011comparing}, we propose to add special tags between words.
We use starting and ending tags before and after words supporting semantic concepts. Starting tags also define the nature of the concept, and there are as many different starting tags as different concepts to extract, while the same ending tag is used to delimit the end of the concept.
For instance, the sentence "I would like two double-bed rooms" is semantically represented as "I would like \textbf{\textless nb\_room} two~\textbf{\textgreater} \textbf{\textless room\_type} double-bed rooms\textbf{\textgreater}", where '\textbf{\textless nb\_room}' and '\textbf{\textless room\_type}' are two starting tags defining two different semantic concepts (number of rooms and room type) while '\textgreater' is the unique symbol to represent the end of a concept. 
Since the neural model emits characters, in practice each starting tag is represented by single symbol instead of a sequence of characters.
Last, notice that in this example 'two' is the value of concept 'number of rooms' while 'double-bed room' is the value of 'room type'.

In order to make the CTC loss function focus more on concepts and their values instead of unlabelled words, we also introduced in~\cite{ghannay2018end} the starred mode.
It consists on replacing all the characters comprise between two semantic concept by a single star. 
The previous example becomes, under starred mode: "*~\textbf{\textless nb\_room} two~\textbf{\textgreater} \textbf{\textless room\_type} double-bed rooms\textbf{\textgreater}".
The goal of this starred approach was to improve semantic extraction by strongly penalizing errors localized on areas of semantic interests during the training process.

\section{Curriculum-based transfer learning}

Intuition of curriculum learning is based on the analogy with humans who learn better when concepts to be learnt and examples are presented gradually, from simple ones to more complex ones.
The motivation of curriculum learning is that the order the training data is presented, from easy examples to more difficult ones, helps training algorithms, for instance by accelerating the  convergence and by guiding the learner towards better local \textsl{minima}~\cite{krueger2009flexible}. 
A curriculum learning strategy can also be considered as a special form of transfer learning where the first learnt tasks are used to guide the learner so that it will perform better on the final task~\cite{bengio2009curriculum}.
In this study, we aim to hybrid both curriculum learning strategy and more classical transfer learning.

\begin{figure}[!htbp]
  \centering
  \includegraphics[width=0.9\columnwidth]{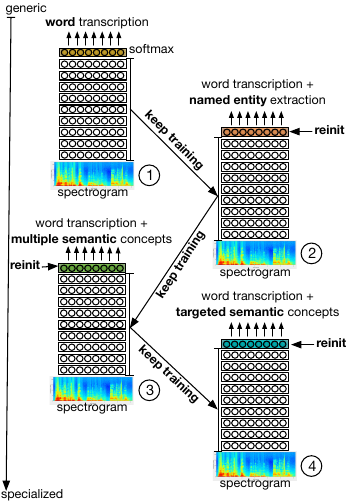}
  \caption{Example of curriculum-based transfer learning for end-to-end spoken language understanding from speech}
  \label{fig:curriculum}
\end{figure}

To train an end-to-end neural model for spoken language understanding that directly takes speech as input, we need both audio recordings and their semantic annotations.
A first remark consists on underlining that such training data are usually limited in size, and are probably not large enough to train an effective SLU system.
A second remark is about the availability of other resources containing both audio recordings and manual annotations. 
For any resourced languages, like English or French, such resources exist and their use must be considered to help to train an SLU end-to-end neural model. 
These resources can be simple audio recordings with manual transcriptions, but can also be audio recordings with manual annotations that express some semantic aspects, not directly related to the final semantic task.
For instance, in French several corpora exist that contain annotations on named entities or semantic concepts for different tasks related to goal-oriented human/machine dialogues.
In order to take benefit from the existence of these data to train an SLU end-to-end system, we suggest to order these data from the most semantically generic to the most specific ones, and to train successive neural models by reinjecting the weights trained at step $t$ as preinitialized weights at step $t+1$, except for the output layer that has to be reinitialized to handle new output symbols. 
Figure~\ref{fig:curriculum} illustrates this approach.

First, we consider the most semantically generic data as the ones containing manual transcriptions (\textsl{cf.} only words) of audio recordings. 
Secondly, we consider the use of audio recordings associated to manual annotations of named entities. 
We assume that named entities recognition and slot filling task based on semantic concept extraction are very close SLU tasks and we assume that named entities are more generic semantic concepts than the semantic concepts designed to specialized human/machine dialogues.
Third, we merge the different semantic concepts designed to different specialized human/machine dialogues into a same set of concepts and, fourth, we only focus on training data of the final targeted task.

This approach is not pure curriculum learning since at each training step the targeted task changes. 
But except for the softmax output layer, all the parameters continue their training step by step. 
Since the output symbols depend on the task, the output layer is reinitialized at each training step. 
The curriculum-based transfer learning approach proposed in this paper consists on designing a sequence of transfer learning processes that follows the principles of curriculum learning: from simple tasks to more complex ones.
\vspace{-0.3cm}
\section{Experiments}
\vspace{-0.3cm}
\subsection{Data}
\label{sec:data}
Experiments were carried out on French corpora that are accessible, making reproducible these experiments. 
Data used for the ASR system training come from five different sources: EPAC \cite{esteve2010epac}, ESTER~2 \cite{galliano2009ester}, ETAPE \cite{gravier2012etape}, QUAERO \cite{grouin2011proposal} and REPERE \cite{giraudel2012repere}. 
These data were recorded from French speaking radio and TV stations between 1998 and 2012. 
All these audio recordings were manually transcribed and divided into three parts: training, development, and evaluation sets. 
Our final data set is the merge of all these corpora respecting the official distribution.

Manual annotations of named entities are available for the ETAPE and QUAERO corpora according to 8 main types: \textit{amount}, \textit{event}, \textit{func}, \textit{loc}, \textit{org}, \textit{pers}, \textit{prod} and \textit{time}. 
We used these manually annotated data to train a state-of-the-art (text-to-text) sequence labelling system\footnote{https://github.com/XuezheMax/NeuroNLP2}. 
Thanks to this system, we automatically annotated all the ASR data that were not initially manually annotated on named entities.
Experiments needing named entities were carried out on the full ASR data set with the combination of manual and automatic named entity annotations.

The slot filling annotated data come from two different sources: MEDIA \cite{bonneau2005semantic} and PORTMEDIA \cite{lefevre2012leveraging}. 
Both are composed of telephone conversations. 
The MEDIA corpus is dedicated to hotel booking. 
It is composed of 1257 dialogues and split into three parts: a training set containing 12.9k sentences, a development set containing 1.3k sentences, and an evaluation set containing 3.5k sentences. 
This corpus is manually annotated with 76 semantics concepts (e.g. "number of rooms", "hotel name", "localization", "room equipment", ...). 
The PORTMEDIA corpus is dedicated to theater tickets reservation. 
It is composed of 700 dialogues and is also split into three parts: a training set containing 5.9k sentences, a development part containing 1.4k sentences, and an evaluation set containing 2.8k sentences. 
PORTMEDIA corpus is manually annotated with 36 semantics concepts close to the MEDIA concept set: PORTMEDIA and MEDIA share 26 common semantic concepts.

\subsection{Performance of curriculum-based transfer learning}


First experiments target the slot filling task in the MEDIA domain.
Table~\ref{tab:greedyMEDIA} presents results in greedy mode on the MEDIA test set, in which we can notice that exploiting only the MEDIA data to train an end-to-end neural model ($SF_M$) for slot filling task leads to a concept error rate (CER) or 39.8\% while the concept/value error rate (CVER) reaches 52.1\%.
CER is a metrics similar to the word error rate metrics but applied on concepts.
CVER is very close to CER but evaluate concept/value pairs instead of evaluating only concepts.
Initializing weights pretrained on the ASR task ($ASR$) before training $SF_M$ very strongly reduces both CER and CVER, since the system $ASR \bullet SF_M$ reaches 23.7\% of CER and 30.3\% of CVER.
Exploiting the merge of the MEDIA and PORT-MEDIA corpora as an intermediate transfer learning task ($SF_{P+M}$) between $ASR$ and $SF_M$ training processes allows us to reduce again both CER and CVER.
Last, the deep neural model trained by  following the complete curriculum-based transfer learning proposed in this paper reaches 21.6\% in greedy mode. This complete chain training integrates the name entity recognition tasks, called $NER$, between $ASR$ and $SF_{P+M}$ training steps.

\begin{table}[th]
  \caption{Performance of end-to-end SLU system in greedy decoding on the MEDIA test}
  \label{tab:greedyMEDIA}
\vspace{-0.3cm}

  \centering
  \begin{tabular}{|l|c|c|}
    \hline    
    Training chain& CER & CVER \\
    \hline
    \hline

    $SF_M$                                     & 39.8 & 52.1\\
    \hline
    $ASR \bullet SF_M$                              & 23.7 & 30.3\\
    \hline
    $ASR \bullet SF_{P+M} \bullet SF_{M}$           & 22.2 & 28.8\\
    \hline
    $ASR \bullet NER \bullet SF_{P+M} \bullet SF_{M}$ & 21.6 & 27.7\\
    \hline
  \end{tabular}
\end{table}

Like for the Deep Speech~2 Baidu's system, it is possible to apply a word-level language model (LM) rescoring through a beam search computed on the neural model outputs. 
By applying a such rescoring with a 5-gram LM trained on the MEDIA train set and on out-domain data (mainly from newspaper articles), results are significantly improved.
They are presented in table~\ref{tab:5gMEDIA}.
As awaited, all results are improved and the curriculum-based transfer learning is still useful, making possible to reach 18.1\% of CER and 22.1\% of CVER.


\begin{table}[th]
  \caption{Performance of end-to-end SLU system with beam-search decoding with a 5-gram LM rescoring on the MEDIA test}
  \label{tab:5gMEDIA}
\vspace{-0.3cm}
  \centering
  \begin{tabular}{| l | c | c |}
\hline    

Training chain & CER & CVER \\
\hline    
\hline    

    $SF_M$                                     & 32.8 & 37.9\\
\hline    
    $ASR \bullet SF_M$                              & 20.1 & 24.0\\
\hline    
    $ASR \bullet SF_{P+M} \bullet SF_{M}$           & 19.0 & 22.9\\
\hline    
    $ASR \bullet NER \bullet SF_{P+M} \bullet SF_{M}$ & 18.1 & 22.1\\
\hline    
  \end{tabular}
\end{table}

To reduce more the CER/CVER values, we use the starred mode we introduced in~\cite{ghannay2018end} and described in section~\ref{sec:neuralarch}.
Table~\ref{tab:results_star_5g} presents the results reached on the MEDIA test set. The $\star$ symbol is used to indicate a neural model emits its outputs though the starred mode. 
Notice that the best results are reached when the two last training processes in the curriculum chain use this mode. 
These training processes, $SF_{P+M}(\star)$ and $SF_{M}(\star)$ use very close semantic annotations as output.
When we also applied the starred mode on the $NER$ training process, we degraded the global results.
We think the starred mode must not be applied too early in the training chain, and would be applied on sub-tasks very close to the final target.
In comparison to the literature, 16.4\% of CER and 20.9\% of CVER are very good results, since the best results ever published on the MEDIA test, when analysing speech instead of manual transcriptions, were a CER of 19.9\% and a CVER of 25.1\%~\cite{simonnet2017asr}.
For fair comparison with a state-of-the-art approach, we present in the next section a pipeline system we develop that takes benefits of the very good quality of our ASR outputs.

\begin{table}[th]
  \caption{Performance of end-to-end SLU system in starred mode with beam-search  decoding on the MEDIA test}
  \label{tab:results_star_5g}
  \centering
  \vspace{-0.3cm}

  \begin{tabular}{|l|c|c|}
    \hline
    Training  chain & CER & CVER \\
    \hline
    \hline
    $ASR \bullet SF_{P+M} \bullet SF_{M}(\star)$             & 17.0 & 21.5\\
    \hline
    $ASR \bullet NER \bullet SF_{P+M}(\star) \bullet SF_{M}$ & 18.0 & 22.0 \\
    \hline
    $ASR \bullet NER \bullet SF_{P+M}(\star) \bullet SF_{M}(\star)$ & 16.4 & 20.9 \\
    \hline
  \end{tabular}
\end{table}


\vspace{-0.3cm}
\subsection{Comparison to state-of-the-art pipeline approach}
Two classical SLU systems are implemented based on a Conditional Random Field (CRF) model\footnote{the Wapiti toolkit is used \cite{lavergne2010practical}}. They only differ by the set of features chosen which is defined in a template. Indeed, the template indicates the set of features the training patterns have to based on for the CRF system to learn the model. For the sake of simplicity, the template indicates which features represent each current word. The following features are available: (i) the word itself (its surface form); (ii) its pre-defined \emph{semantic} categories belonging to MEDIA specific categories like names of the streets, cities or hotels, lists of room equipments, food type... (\textit{e.g.} TOWN for Paris), or more general categories like figures, days, months... (\textit{e.g.} FIGURE for thirty-three); (iii) a set of \emph{syntactic} features extracted by the MACAON toolkit \cite{nasr2010macaon}. As a result we obtain for each word its lemma, POS tag, word governor and its relation with the current word; (iv) a set of \emph{morphological} features corresponding to the 1-to-3 first letter ngrams, and the 1-to-3 last letter ngrams of the word.

In order to evaluate the contribution of semantic, syntactic and morphological features, we choose to design two templates: one considering only the surface form of the word, and the other one considering all available features. Systems based on these two templates are respectively denoted $SF^{CRF}_{lex}$ and $SF^{CRF}_{lex+feat}$. These systems process slot filling on automatic transcriptions. For our experiments, we feed them with outputs from our end-to-end $ASR$ system (with LM rescoring) fine tuned one the MEDIA training data. 
The word error rate (WER) of this system on the MEDIA data is 9.3\%, that is very good in comparison to the 23.6\% of WER got by the ASR used to reach the best CER/CVER values in the literacy until now~\cite{simonnet2017asr}.
Comparisons in CER/CVER values between state-of-the art pipeline approach and end-to-end system are provided in table~\ref{tab:results_stars_5g_pipeline}.
The pipeline approach reaches a CER of 16.1\% and a CVER of 20.9\% that is slightly better than the results reached by the end-to-end approach. 
By computing the 95\% confidence interval through the Student's t-test, we can observe than the confidence margin is 0.7 for CER (0.8 for CVER). 
By the way, this means that the differences between the pipeline and the end-to-end approaches are not statistically significant.
Another remark comes from the results got with $SF^{CRF}_{lex}$: this system uses only lexical information and can be directly compared to the end-to-end approach that actually does not use external features coming from human expertise or NLP tools like predefined semantic categorie, POS, word governor and dependencies... 
In this case, the comparison is largely at the advantage of the end-to-end system. 
This shows that it would be important to investigate in the future how to inject to the end-to-end model external information that help so much in the pipeline approach.

\begin{table}[th]
  \caption{Comparison between state-of-the art pipeline approach and end-to-end  system on the MEDIA test }
  \label{tab:results_stars_5g_pipeline}
  \centering
  \vspace{-0.3cm}
  \begin{tabular}{|l|c|c|}
    \hline
    System & CER & CVER \\
    \hline
    \hline
    $Pip[ASR_M\rightarrow SF^{CRF}_{lex}]$ & 20.6 & 24.8\\
    \hline
    $Pip[ASR_M\rightarrow NLP \rightarrow SF^{CRF}_{lex+feat}]$ & 16.1 & 20.4\\
    \hline
    $ASR \bullet NER \bullet SF_{P+M}(\star) \bullet SF_{M}(\star)$ & 16.4 & 20.9\\
    \hline
  \end{tabular}
\end{table}

\vspace{-0.3cm}
\subsection{Domain portability}
Last experiments address the portability issue for SLU systems. The PORTMEDIA corpus was produced in order to investigate two axes of portability: domain and language~\cite{lefevre2012leveraging}. In this work we only focus on domain portability by using the PORTMEDIA corpus previously described, that is actually only a sub part of the PORTMEDIA data that also contain an Italian version of MEDIA. 
As seen in section~\ref{sec:data}, PORTMEDIA training data contains twice less data than the MEDIA training.
Table~\ref{tab:results_portmedia} presents several results thanks to different training chains. $ASR \bullet SF_{M} \bullet SF_{P}$ shows that we can reach 25.2\% of CER when the $SF_{P}$ model is trained from a system dedicated to MEDIA ($ASR \bullet SF_{M}$), that is better than starting directly from the $ASR$ model or from scratch.
At the end, to reach the best result we can, the approach consists on applying the  same curriculum-based transfer learning used in the previous experiments when targeting  MEDIA.
In order to reduce computational costs to develop an SLU dedicated to a new task, it seems that the best way is to save the $ASR \bullet NER$ model and to start the new training chain from it. 
$ASR \bullet NER$ seems to be a very relevant starting point to tackle slot filling tasks. Notice that around 4 weeks of computation are needed to train the $ASR \bullet NER$ model on two Titan X (Pascal) GPU cards, while only a few days are need for the $SF_{P+M}(\star) \bullet SF_{P}(\star)$ part.

\begin{table}[th]
  \caption{Results on the PORTMEDIA test data for domain portability}
  \label{tab:results_portmedia}
  \centering
  \vspace{-0.3cm}
  \begin{tabular}{|l|c|c|}
    \hline
    Training chain & CER & CVER \\
    \hline
    \hline
    $SF_P$                                 & 42.3 & 57.9\\
    \hline
    $ASR \bullet SF_{P}$                        & 27.4 & 40.0\\
    \hline
    $ASR \bullet NER \bullet SF_{P}$            & 26.8 & 38.9\\
    \hline
    $ASR \bullet SF_{M} \bullet SF_{P}$         & 25.2 & 37.5\\
    \hline
    $ASR \bullet SF_{P+M} \bullet SF_{P}$       & 24.9 & 36.7\\
    \hline
    $ASR \bullet NER\bullet SF_{P+M} \bullet SF_{P}$ & 24.1 & 35.9\\
    \hline
    $ASR \bullet NER \bullet SF_{P+M} \bullet SF_{P}(\star)$ & 22.3 & 36.5\\
    \hline
    $ASR \bullet NER \bullet SF_{P+M}(\star) \bullet SF_{P}(\star)$ & 21.9 & 36.9\\
    \hline
  \end{tabular}
\end{table}

\vspace{-0.3cm}
\section{Conclusion}
We propose a curriculum-based transfer learning approach that allows us to train a very competitive SLU end-to-end system from speech that gets state-of-the-art results.
This approach can also be applied in order to train a model dedicated to a new slot filling task from an already pre-trained model (here $ASR \bullet NER$), in the same spirit as the BERT model for textual language understanding~\cite{devlin2018bert}.

We think we will outperform soon the current state-of-art approach by injecting external information. For instance, our current investigations on speaker adaptation and language transfer for the MEDIA task, not presented in this paper by lack of space, also provide very competitive and complementary results~\cite{adaptation2019}.

\section{Acknowledgements}
This work was supported by the French ANR Agency through the CHIST-ERA ON-TRAC project, under the contract number ANR-18-CE23-0021-01, and by the RFI Atlanstic2020 RAPACE project.

\bibliographystyle{IEEEtran}

\bibliography{mybib}

\begin{thebibliography}{10}
\providecommand{\url}[1]{#1}
\csname url@samestyle\endcsname
\providecommand{\newblock}{\relax}
\providecommand{\bibinfo}[2]{#2}
\providecommand{\BIBentrySTDinterwordspacing}{\spaceskip=0pt\relax}
\providecommand{\BIBentryALTinterwordstretchfactor}{4}
\providecommand{\BIBentryALTinterwordspacing}{\spaceskip=\fontdimen2\font plus
\BIBentryALTinterwordstretchfactor\fontdimen3\font minus
  \fontdimen4\font\relax}
\providecommand{\BIBforeignlanguage}[2]{{%
\expandafter\ifx\csname l@#1\endcsname\relax
\typeout{** WARNING: IEEEtran.bst: No hyphenation pattern has been}%
\typeout{** loaded for the language `#1'. Using the pattern for}%
\typeout{** the default language instead.}%
\else
\language=\csname l@#1\endcsname
\fi
#2}}
\providecommand{\BIBdecl}{\relax}
\BIBdecl

\bibitem{kubala1998named}
F.~Kubala, R.~Schwartz, R.~Stone, and R.~Weischedel, ``Named entity extraction
  from speech,'' in \emph{Proceedings of DARPA Broadcast News Transcription and
  Understanding Workshop}.\hskip 1em plus 0.5em minus 0.4em\relax Citeseer,
  1998, pp. 287--292.

\bibitem{gorin1997may}
A.~L. Gorin, G.~Riccardi, and J.~H. Wright, ``How may i help you?''
  \emph{Speech communication}, vol.~23, no. 1-2, pp. 113--127, 1997.

\bibitem{yaman2008integrative}
S.~Yaman, L.~Deng, D.~Yu, Y.-Y. Wang, and A.~Acero, ``An integrative and
  discriminative technique for spoken utterance classification,'' \emph{IEEE
  Transactions on Audio, Speech, and Language Processing}, vol.~16, no.~6, pp.
  1207--1214, 2008.

\bibitem{tur2012towards}
G.~Tur, L.~Deng, D.~Hakkani-T{\"u}r, and X.~He, ``Towards deeper understanding:
  Deep convex networks for semantic utterance classification,'' in \emph{2012
  IEEE international conference on acoustics, speech and signal processing
  (ICASSP)}.\hskip 1em plus 0.5em minus 0.4em\relax IEEE, 2012, pp. 5045--5048.

\bibitem{morchid2013theme}
M.~Morchid, G.~Linares, M.~El-Beze, and R.~De~Mori, ``Theme identification in
  telephone service conversations using quaternions of speech features.'' in
  \emph{INTERSPEECH}, 2013, pp. 1394--1398.

\bibitem{tur2011spoken}
G.~Tur and R.~De~Mori, \emph{Spoken language understanding: Systems for
  extracting semantic information from speech}.\hskip 1em plus 0.5em minus
  0.4em\relax John Wiley \& Sons, 2011.

\bibitem{chen2013unsupervised}
Y.-N. Chen, W.~Y. Wang, and A.~I. Rudnicky, ``Unsupervised induction and
  filling of semantic slots for spoken dialogue systems using frame-semantic
  parsing,'' in \emph{2013 IEEE Workshop on Automatic Speech Recognition and
  Understanding}.\hskip 1em plus 0.5em minus 0.4em\relax IEEE, 2013, pp.
  120--125.

\bibitem{simonnet2017asr}
E.~Simonnet, S.~Ghannay, N.~Camelin, Y.~Est{\`e}ve, and R.~De~Mori, ``{ASR}
  error management for improving spoken language understanding,'' in
  \emph{Interspeech 2017}, 2017.

\bibitem{ghannay2018end}
S.~Ghannay, A.~Caubri{\`e}re, Y.~Est{\`e}ve, N.~Camelin, E.~Simonnet,
  A.~Laurent, and E.~Morin, ``End-to-end named entity and semantic concept
  extraction from speech,'' in \emph{2018 IEEE Spoken Language Technology
  Workshop (SLT)}.\hskip 1em plus 0.5em minus 0.4em\relax IEEE, 2018, pp.
  692--699.

\bibitem{bengio2009curriculum}
Y.~Bengio, J.~Louradour, R.~Collobert, and J.~Weston, ``Curriculum learning,''
  in \emph{Proceedings of the 26th annual international conference on machine
  learning}.\hskip 1em plus 0.5em minus 0.4em\relax ACM, 2009, pp. 41--48.

\bibitem{bengio2011deep}
Y.~Bengio, ``Deep learning of representations for unsupervised and transfer
  learning,'' in \emph{Proceedings of the 2011 International Conference on
  Unsupervised and Transfer Learning workshop-Volume 27}.\hskip 1em plus 0.5em
  minus 0.4em\relax JMLR. org, 2011, pp. 17--37.

\bibitem{bonneau2005semantic}
H.~Bonneau-Maynard, S.~Rosset, C.~Ayache, A.~Kuhn, and D.~Mostefa, ``Semantic
  annotation of the {French MEDIA} dialog corpus,'' in \emph{Ninth European
  Conference on Speech Communication and Technology}, 2005.

\bibitem{lefevre2012leveraging}
F.~Lef\`evre, D.~Mostefa, L.~Besacier, Y.~Est\`eve, M.~Quignard, N.~Camelin,
  B.~Favre, B.~Jabaian, and L.~M.~R. Barahona, ``Leveraging study of robustness
  and portability of spoken language understanding systems across languages and
  domains: the {PORTMEDIA} corpora,'' in \emph{The International Conference on
  Language Resources and Evaluation}, 2012.

\bibitem{hannun2014deep}
A.~Hannun, C.~Case, J.~Casper, B.~Catanzaro, G.~Diamos, E.~Elsen, R.~Prenger,
  S.~Satheesh, S.~Sengupta, A.~Coates \emph{et~al.}, ``Deep speech: Scaling up
  end-to-end speech recognition,'' \emph{arXiv preprint arXiv:1412.5567}, 2014.

\bibitem{berard2016listen}
A.~B{\'e}rard, O.~Pietquin, L.~Besacier, and C.~Servan, ``Listen and translate:
  A proof of concept for end-to-end speech-to-text translation,'' in \emph{NIPS
  Workshop on end-to-end learning for speech and audio processing}, 2016.

\bibitem{weiss2017sequence}
R.~J. Weiss, J.~Chorowski, N.~Jaitly, Y.~Wu, and Z.~Chen,
  ``Sequence-to-sequence models can directly translate foreign speech,''
  \emph{Proc. Interspeech 2017}, pp. 2625--2629, 2017.

\bibitem{berard2018end}
A.~B{\'e}rard, L.~Besacier, A.~C. Kocabiyikoglu, and O.~Pietquin, ``End-to-end
  automatic speech translation of audiobooks,'' in \emph{2018 IEEE
  International Conference on Acoustics, Speech and Signal Processing
  (ICASSP)}.\hskip 1em plus 0.5em minus 0.4em\relax IEEE, 2018, pp. 6224--6228.

\bibitem{jan2018iwslt}
N.~Jan, R.~Cattoni, S.~Sebastian, M.~Cettolo, M.~Turchi, and M.~Federico, ``The
  iwslt 2018 evaluation campaign,'' in \emph{International Workshop on Spoken
  Language Translation}, 2018, pp. 2--6.

\bibitem{serdyuk2018towards}
D.~Serdyuk, Y.~Wang, C.~Fuegen, A.~Kumar, B.~Liu, and Y.~Bengio, ``Towards
  end-to-end spoken language understanding,'' in \emph{2018 IEEE International
  Conference on Acoustics, Speech and Signal Processing (ICASSP)}.\hskip 1em
  plus 0.5em minus 0.4em\relax IEEE, 2018, pp. 5754--5758.

\bibitem{elman1993learning}
J.~L. Elman, ``Learning and development in neural networks: The importance of
  starting small,'' \emph{Cognition}, vol.~48, no.~1, pp. 71--99, 1993.

\bibitem{platanios2019competence}
E.~A. Platanios, O.~Stretcu, G.~Neubig, B.~Poczos, and T.~M. Mitchell,
  ``Competence-based curriculum learning for neural machine translation,''
  \emph{arXiv preprint arXiv:1903.09848}, 2019.

\bibitem{jabaian2012portability}
B.~Jabaian, F.~Lef{\`e}vre, and L.~Besacier, ``Portability of semantic
  annotations for fast development of dialogue corpora,'' in \emph{Interspeech
  2012}, 2012, p.~xx.

\bibitem{amodei2016deep}
D.~Amodei, S.~Ananthanarayanan, R.~Anubhai, J.~Bai, E.~Battenberg, C.~Case,
  J.~Casper, B.~Catanzaro, Q.~Cheng, G.~Chen \emph{et~al.}, ``Deep speech 2:
  End-to-end speech recognition in english and mandarin,'' in
  \emph{International conference on machine learning}, 2016, pp. 173--182.

\bibitem{graves2006connectionist}
A.~Graves, S.~Fern{\'a}ndez, F.~Gomez, and J.~Schmidhuber, ``Connectionist
  temporal classification: labelling unsegmented sequence data with recurrent
  neural networks,'' in \emph{Proceedings of the 23rd international conference
  on Machine learning}.\hskip 1em plus 0.5em minus 0.4em\relax ACM, 2006, pp.
  369--376.

\bibitem{hahn2011comparing}
S.~Hahn, M.~Dinarelli, C.~Raymond, F.~Lefevre, P.~Lehnen, R.~De~Mori,
  A.~Moschitti, H.~Ney, and G.~Riccardi, ``Comparing stochastic approaches to
  spoken language understanding in multiple languages,'' \emph{IEEE
  Transactions on Audio, Speech, and Language Processing}, vol.~19, no.~6, pp.
  1569--1583, 2011.

\bibitem{krueger2009flexible}
K.~A. Krueger and P.~Dayan, ``Flexible shaping: How learning in small steps
  helps,'' \emph{Cognition}, vol. 110, no.~3, pp. 380--394, 2009.

\bibitem{esteve2010epac}
Y.~Est\`eve, T.~Bazillon, J.-Y. Antoine, F.~B{\'e}chet, and J.~Farinas, ``The
  {EPAC} corpus: Manual and automatic annotations of conversational speech in
  {French} broadcast news.'' in \emph{LREC}, 2010.

\bibitem{galliano2009ester}
S.~Galliano, G.~Gravier, and L.~Chaubard, ``The {ESTER~2} evaluation campaign
  for the rich transcription of {French} radio broadcasts,'' in \emph{Tenth
  Annual Conference of the International Speech Communication Association},
  2009.

\bibitem{gravier2012etape}
G.~Gravier, G.~Adda, N.~Paulson, M.~Carr{\'e}, A.~Giraudel, and O.~Galibert,
  ``The {ETAPE} corpus for the evaluation of speech-based {TV} content
  processing in the {French} language,'' in \emph{LREC-Eighth international
  conference on Language Resources and Evaluation}, 2012, p.~na.

\bibitem{grouin2011proposal}
C.~Grouin, S.~Rosset, P.~Zweigenbaum, K.~Fort, O.~Galibert, and L.~Quintard,
  ``Proposal for an extension of traditional named entities: From guidelines to
  evaluation, an overview,'' in \emph{Proceedings of the 5th Linguistic
  Annotation Workshop}.\hskip 1em plus 0.5em minus 0.4em\relax Association for
  Computational Linguistics, 2011, pp. 92--100.

\bibitem{giraudel2012repere}
A.~Giraudel, M.~Carr{\'e}, V.~Mapelli, J.~Kahn, O.~Galibert, and L.~Quintard,
  ``The {REPERE} corpus: a multimodal corpus for person recognition.'' in
  \emph{LREC}, 2012, pp. 1102--1107.

\bibitem{lavergne2010practical}
T.~Lavergne, O.~Capp{\'e}, and F.~Yvon, ``Practical very large scale crfs,'' in
  \emph{Proceedings of the 48th Annual Meeting of the Association for
  Computational Linguistics}.\hskip 1em plus 0.5em minus 0.4em\relax
  Association for Computational Linguistics, 2010, pp. 504--513.

\bibitem{nasr2010macaon}
A.~Nasr, F.~B{\'e}chet, and J.-F. Rey, ``Macaon: Une cha{\^\i}ne linguistique
  pour le traitement de graphes de mots,'' in \emph{Traitement Automatique des
  Langues Naturelles}, 2010.

\bibitem{devlin2018bert}
J.~Devlin, M.-W. Chang, K.~Lee, and K.~Toutanova, ``Bert: Pre-training of deep
  bidirectional transformers for language understanding,'' \emph{arXiv preprint
  arXiv:1810.04805}, 2018.

\bibitem{adaptation2019}
N.~Tomashenko, A.~Caubri\`ere, and Y.~Est{\`e}ve, ``Investigating adaptation
  and transfer learning for end-to-end spoken language understanding from
  speech,'' in \emph{Submitted to Interspeech 2019}, 2019.

\end{thebibliography}


\end{document}